\documentclass[acmtog, nonacm]{acmart}
\acmSubmissionID{}

\usepackage{xcolor}
\definecolor{red}{RGB}{192,0,0}
\definecolor{green}{RGB}{84,130,53}
\definecolor{blue}{RGB}{47,85,151}

\usepackage[ruled]{algorithm2e} 

\SetAlFnt{\small}
\SetAlCapFnt{\small}
\SetAlCapNameFnt{\small}
\SetAlCapHSkip{0pt}

\acmJournal{TOG}

\begin{document}
\title{Zero-Shot Image Harmonization with Generative Model Prior}

\author{Jianqi Chen}
\author{Yilan Zhang}
\author{Zhengxia Zou}
\author{Keyan Chen}
\author{Zhenwei Shi}
\affiliation{
\institution{Beihang University}
\country{China}}

\begin{teaserfigure}
    \centering
    \includegraphics[width=\textwidth]{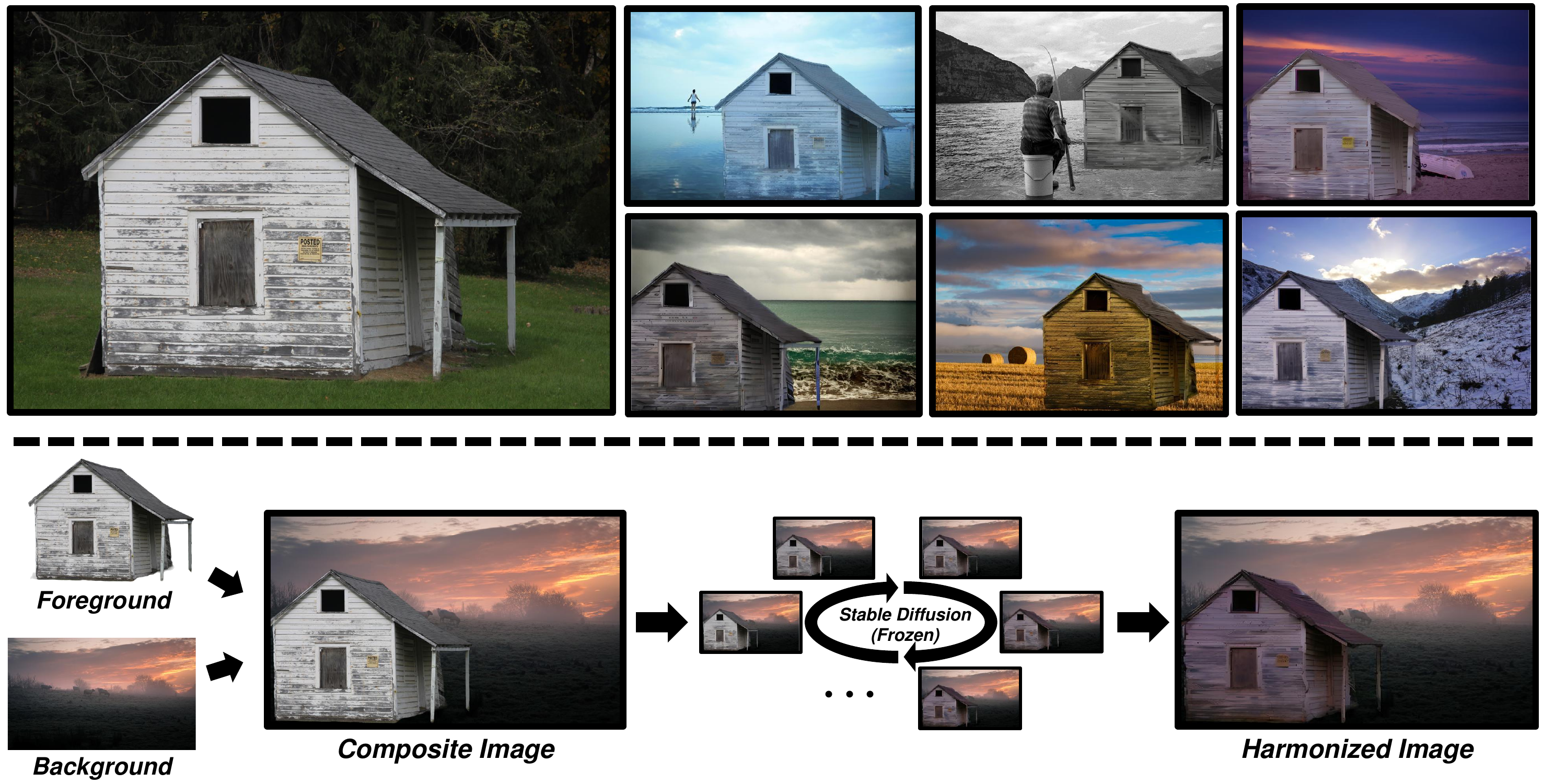}
    \vspace{-2.5em}
    \captionof{figure}{Given a composite image,  our method can achieve its harmonized result, where the color space of the foreground is aligned with that of the background. Our method does not need to collect a large number of composite images for training, but only utilizes pretrained generative models. The first column from the left in the upper row is the source image of the foreground (``house"), and the others are the harmonized results of the foreground object in different backgrounds. In the lower row, we take one of the composite images as an example to show the harmonization process. For a concise overview of our approach, please refer to our \textbf{presentation video: \url{https://www.youtube.com/watch?v=mfBTIVp6JBU&t=4s}}.}
    \label{fig:Preface}
\end{teaserfigure}%

\begin{abstract}
We propose a zero-shot approach to image harmonization, aiming to overcome the reliance on large amounts of synthetic composite images in existing methods. These methods, while showing promising results, involve significant training expenses and often struggle with generalization to unseen images. To this end, we introduce a fully modularized framework inspired by human behavior. Leveraging the reasoning capabilities of recent foundation models in language and vision, our approach comprises three main stages. Initially, we employ a pretrained vision-language model (VLM) to generate descriptions for the composite image. Subsequently, these descriptions guide the foreground harmonization direction of a text-to-image generative model (T2I). We refine text embeddings for enhanced representation of imaging conditions and employ self-attention and edge maps for structure preservation. Following each harmonization iteration, an evaluator determines whether to conclude or modify the harmonization direction. The resulting framework, mirroring human behavior, achieves harmonious results without the need for extensive training. We present compelling visual results across diverse scenes and objects, along with a user study validating the effectiveness of our approach. 

\textbf{Github Code: ~\url{https://github.com/WindVChen/Diff-Harmonization}}
\end{abstract}

\makeatletter
\let\@authorsaddresses\@empty
\makeatother

\maketitle

\section{Introduction}
\label{sec:Introduction}

Image harmonization is a technique for aligning the color space of foreground objects with that of the background in a composite image. Recent state-of-the-art methods \cite{liang2022spatial, xue2022dccf, ke2022harmonizer, cong2022high, guo2022transformer, zhan2020adversarial, sofiiuk2021foreground, chen2023dense} are mostly based on deep learning networks which have achieved more promising results than traditional ones \cite{lalonde2007using, xue2012understanding, reinhard2001color, pitie2005n, sunkavalli2010multi}. However, the performance of these methods, either performing image harmonization in a supervised manner with paired inharmonious/harmonized images \cite{xue2022dccf, ke2022harmonizer, hang2022scs, cong2022high, jiang2021ssh, cong2020dovenet, tsai2017deep, guo2022transformer, chen2023dense} or constructing a zero-sum game by means of GAN \cite{NIPS2014_5ca3e9b1} technique \cite{zhan2020adversarial, chen2019toward}, is highly correlated with the quality of the collection of composite images. If the collected composite images could not cover real-world situations, then the trained network will fail to obtain satisfactory results in practical use. 

Compared to curating a collection of natural images, creating a real-world composite image entails higher labor costs. This involves extracting a foreground object from one image and seamlessly integrating it into another at an appropriate location. To mitigate the impractical cost associated with constructing a large real composite image dataset, prevailing methods \cite{tsai2017deep, chen2019toward, cong2020dovenet} opt for the ``synthesis" of composite images. This is achieved by applying diverse color transformations \cite{reinhard2001color, xiao2006color, pitie2007automated, fecker2008histogram, lee2016automatic} to the semantic regions of extensive segmentation datasets \cite{bychkovsky2011learning, lin2014microsoft, zhou2019semantic}. These transformations yield ``synthesized" composite images along with their paired harmonized counterparts (\textit{i.e.}, the originals). While incorporating more transformations improves proximity to the distribution of real-world composite images, it also results in a substantial dataset \cite{cong2020dovenet}, imposing a heavy burden during training. In this paper, we contend that such a resource-intensive approach diverges from typical human behavior.

\begin{figure}[t]
  \centering
   \includegraphics[width=0.9\linewidth]{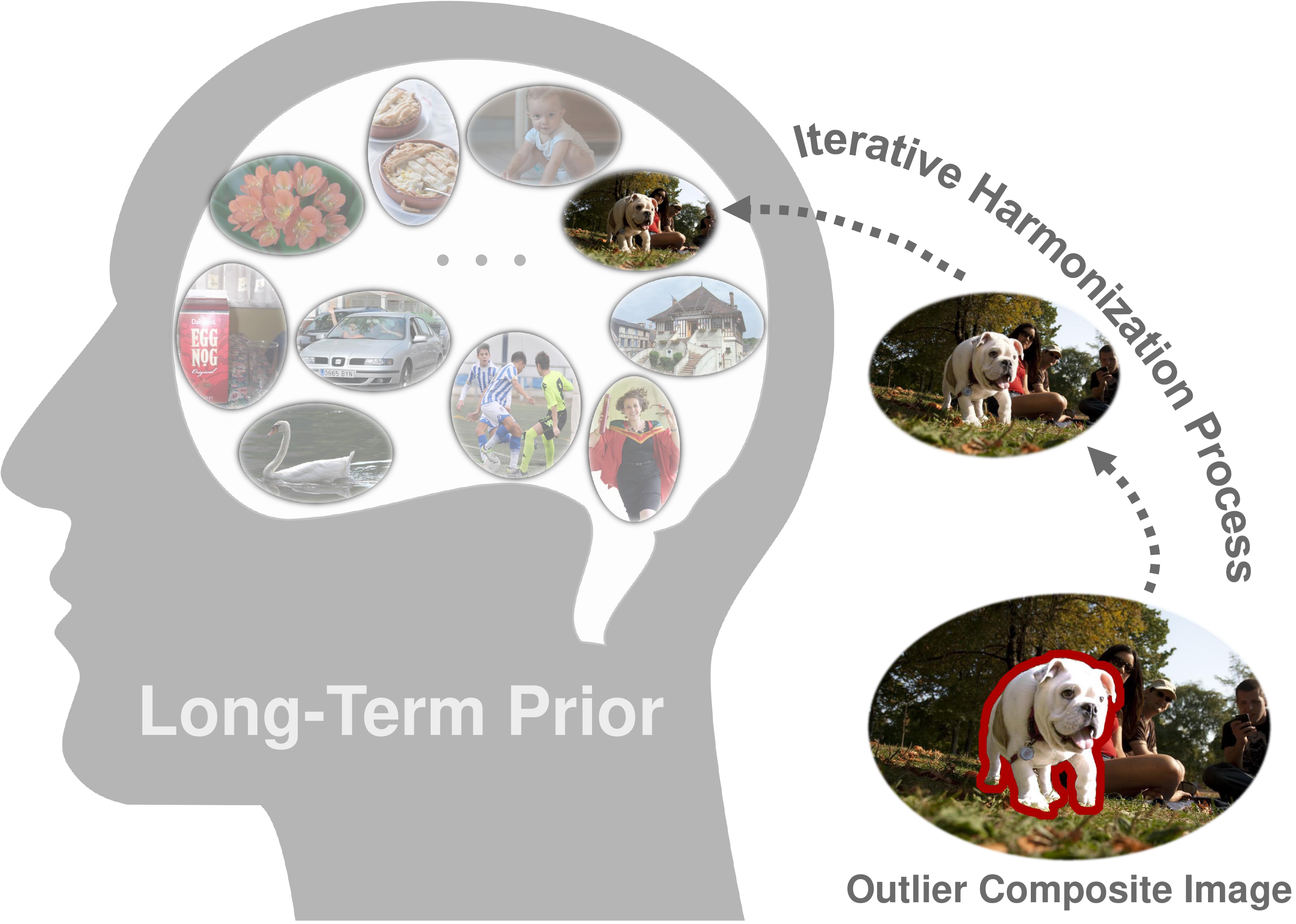}
   \vspace{-2ex}
   \caption{Human behavior of image harmonization. We humans can perform harmonization relying only on our long-term prior, without seeing many composite images in advance. \textit{E.g.}, to harmonize the overbright dog above.}
   \label{fig:HumanBehavior}
   \vspace{-4ex}
\end{figure}

Let's consider how humans harmonize a composite image. 1) First, we quickly identify the possible imaging environment of the foreground and the background regions, including factors such as time of day, season, and added color tones. 2) Next, we mentally visualize the foreground's appearance in the background's imaging condition and utilize basic controls in image editing software, such as Photoshop, to edit the foreground accordingly. 3) After each round of editing, we pause to evaluate the result, determining whether it has achieved harmonization. If successful, we conclude the process; otherwise, we consider refining the current direction. This process doesn't necessitate human training on large datasets of inharmonious images and operates in a zero-shot manner. We attribute its success and feasibility to humans' strong prior understanding of what a natural image should entail. Specifically, most of our observations in our lives are of natural and harmonious scenes, influencing our tendency to imagine and create harmonious compositions. Therefore, the harmonization process is to gradually bring an initial outlier (inharmonious composite image) closer to our established prior (illustrated in Fig. \ref{fig:HumanBehavior}). There is no need to understand the distribution of composite images. This concept finds strong support in early human cognitive research \cite{biederman1982scene} and \cite{davenport2004scene}, which demonstrated humans' rapid and accurate perception of inconsistencies between objects and backgrounds in scenes. This underscores humans' innate ability to harmonize visual elements, forming the theoretical foundation of our approach. 

To emulate human behavior, we identify four key components in the human harmonization process: human reasoning (identifying imaging conditions for foreground and background), human prior regarding natural image appearances, image editing software for operations, and continual intermediate evaluation for adjustments. Drawing inspiration from the advancements in foundation models \cite{bommasani2021opportunities}, we propose a comprehensive solution. For human reasoning, we leverage a vision-language model (VLM) to generate initial descriptions of the imaging condition of composite images. To incorporate human priors, we rely on large text-to-image (T2I) generative models trained on extensive natural image datasets, embedding intrinsic knowledge of natural image distribution. We employ text-guided editing technologies as editing tools and use VLM-rendered descriptions as guidance. Inspired by textual inversion technique \cite{gal2022image}, text embeddings are optimized for a more accurate representation of imaging conditions, and we adopt self-attention maps and edge detection algorithms to ensure structure preservation during editing. For intermediate evaluation, we train a lightweight classifier to assess harmonization levels. The final framework strictly follows human behavior, achieving satisfactory harmonized results. Fig. \ref{fig:Preface} displays some of our results. In summary, our main contributions are:

\begin{itemize}
    \item We present a novel perspective on image harmonization aligned with human behavior, exploring image harmonization through a human-centric lens and assessing its feasibility with current technologies.
    \item We introduce a framework consistent with human behavior, leveraging off-the-shelf VLM and T2I models to achieve satisfactory results without the need for extensive composite image collections. Our designs also handle description inaccuracies, structure distortion, and editing deviation.
    \item Through experiments across diverse cases, we showcase our method's visual effectiveness. Comparative user studies with heavily supervised methods underscore the potential of our zero-shot harmonization approach.
\end{itemize}

\begin{figure*}[t]
  \centering
   \includegraphics[width=\textwidth]{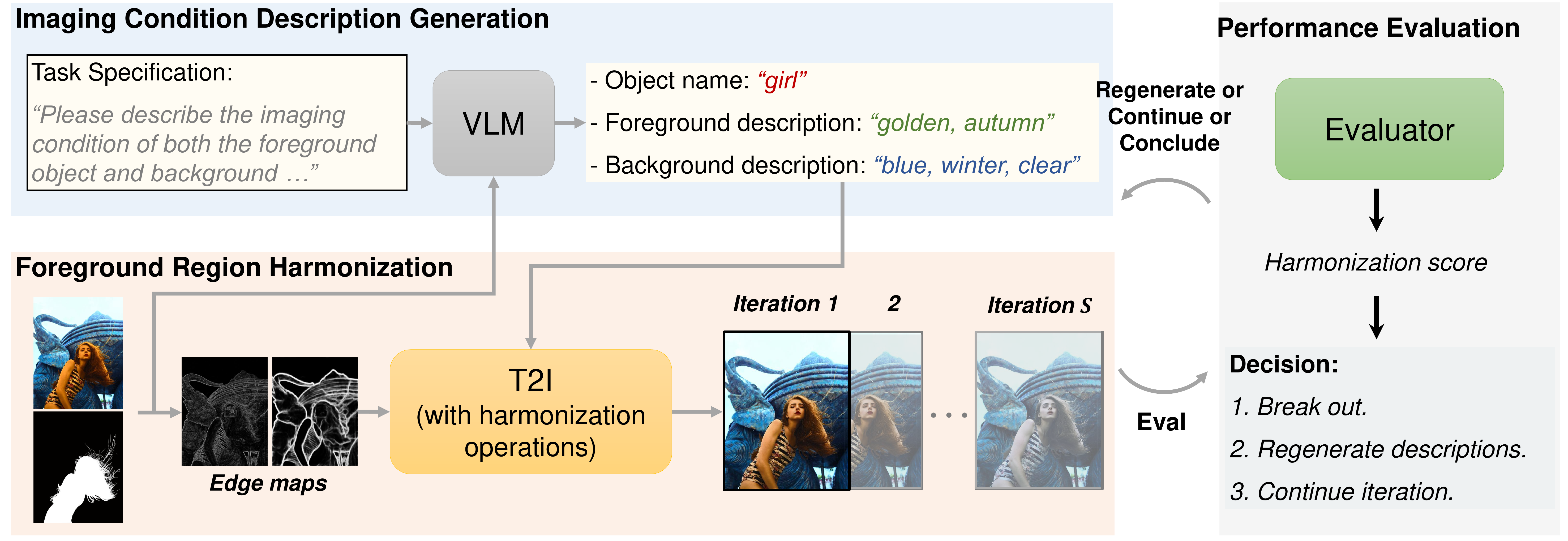}
   \vspace{-6ex}
   \caption{The proposed framework and workflow across modules. The framework comprises three main components: (a) Imaging Condition Description Generation: A vision-language model (VLM) is utilized to generate descriptions of the input composite image, detailing \textit{what the object is} and \textit{how the foreground and background region are}. (b) Foreground Region Harmonization: A text-to-image (T2I) diffusion generative model is employed, taking both the previously generated descriptions and the edge map of the composite images as input. Image editing technology, together with specific harmonization operations, is used to achieve image harmonization. (c) Performance Evaluation: A two-class classifier serves as an evaluator to determine whether the current result is natural-looking enough or if the description should be regenerated, or if the harmonization iteration should continue.}
   \label{fig:Pipeline}
   \vspace{-3ex}
\end{figure*}

\section{Related Work}
\label{sec:relate}

\textbf{Image harmonization}. Compared with traditional methods \cite{lalonde2007using, xue2012understanding, pitie2005n, sunkavalli2010multi} that rely on matching low-level statistics between foreground and background, Tsai \textit{et al.} \cite{tsai2017deep} pioneered leveraging deep learning in image harmonization and demonstrated the superiority of its powerful semantic representation capability. To suffice the data-hungry training of deep networks, Cong \textit{et al.} \cite{cong2020dovenet} constructed a large synthesized image harmonization dataset, by applying a variety of color transforms \cite{reinhard2001color, xiao2006color, pitie2007automated, fecker2008histogram, lee2016automatic} on segmentation regions of the existing datasets \cite{bychkovsky2011learning, lin2014microsoft, laffont2014transient}. Many of the following works \cite{xue2022dccf, ke2022harmonizer, cong2022high, guo2022transformer, sofiiuk2021foreground, chen2023dense} then based their methods on this large synthesized dataset. Some other approaches have either tried to leverage different transforms \cite{hang2022scs, jiang2021ssh, zhang2023controlcom} or leverage the GAN \cite{NIPS2014_5ca3e9b1} technique to explore harmonization with unpaired image data \cite{chen2019toward, zhan2020adversarial, wang2023semi}, but still heavily rely on transforming the segmented regions for composite image collection. Although some promising results have been achieved, whether the synthesized composite images can reflect the real-world situation determines the final generalization and performance of these methods. With more color transformations considered, the trained network can be more robust, yet the training cost also be more unaffordable.

\textbf{Text-to-image synthesis}. Recent advancements in large-scale text-to-image models showcase remarkable generative capabilities. Trained on extensive image-text datasets, these models inherently capture the natural data distribution, enabling them to generate stunning images that align well with given textual guidance. Notably, diffusion models \cite{saharia2022photorealistic, ramesh2022hierarchical, rombach2022high} have garnered significant attention, leading to their adoption in various downstream applications such as image restoration \cite{wang2022zero} and adversarial attacks \cite{chen2023diffusion}. Image editing remains a prevalent application, where these models, guided by text \cite{ho2021classifier}, trained classifiers \cite{dhariwal2021diffusion}, and others \cite{hertz2022prompt, avrahami2022blended}, excel in manipulating local content while preserving the surrounding area. However, these methods are not well-suited for image harmonization, as their primary focus is on altering local content structures.

\section{Approach}
\label{sec:method}

Our method is a fully modularized framework that decomposes the harmonization task into imaging condition description generation, foreground region harmonization, and performance evaluation, as depicted in Fig. \ref{fig:Pipeline}. Given a composite image, we first generate the description of the imaging condition of the foreground and the background with a vision-language model (VLM). Then, a foreground region harmonization module takes these descriptions, along with additional edge maps, to harmonize the foreground region through a large text-to-image (T2I) model. The harmonization result is then assessed by an evaluator to make sure the harmonization direction is correct, or the direction gets revised.

\begin{figure*}[t]
  \centering
   \includegraphics[width=\textwidth]{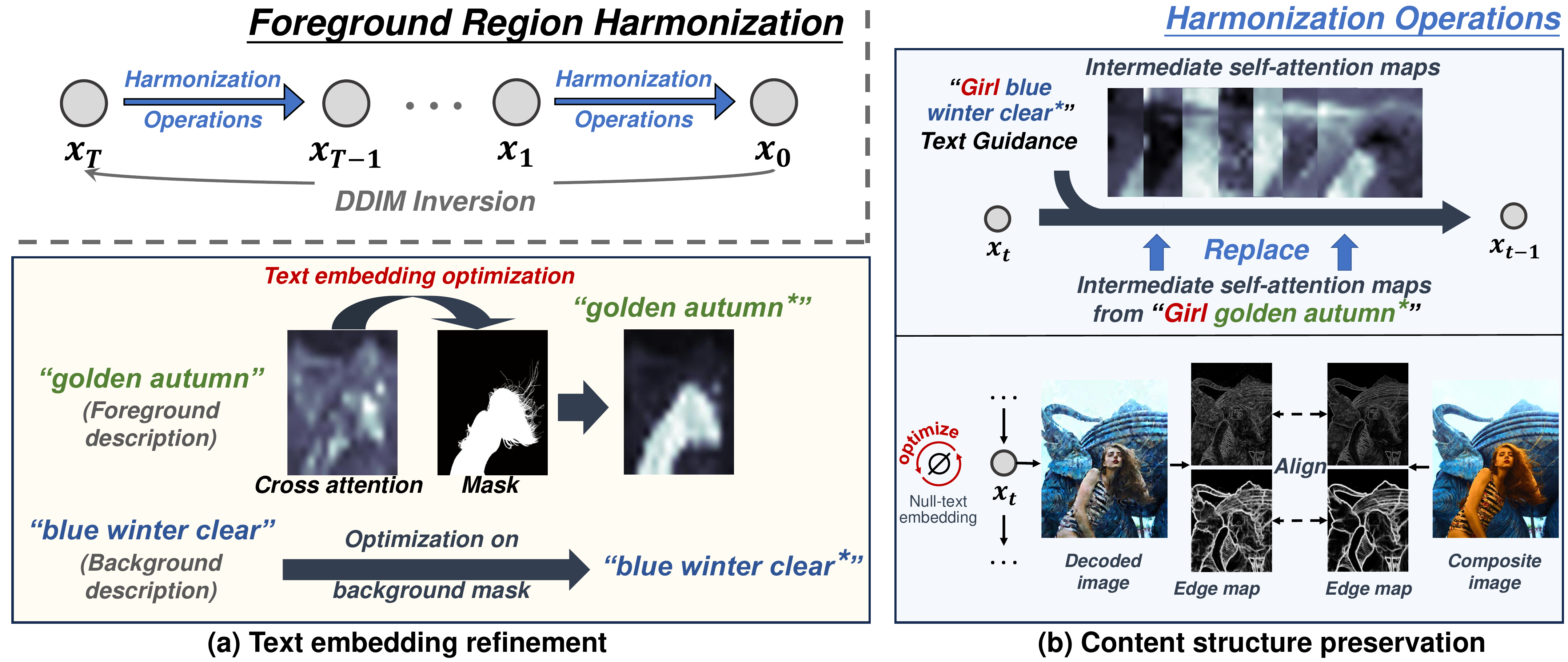}
   \vspace{-6ex}
   \caption{Designs in the Foreground Region Harmonization. Starting from the composite image, we invert the last harmonized result into its diffusion latent and then employ harmonization operations to obtain the next harmonized result. Among these harmonization operations, (a) Text Embedding Refinement is designed to obtain text embeddings that can better represent the foreground/background environment. (b) In Content Structure Preservation, we leverage self-attention maps to retain high-level structure and utilize edge maps to preserve low-level details. The editing process is achieved based on Prompt-to-Prompt (P2P) editing technology \cite{hertz2022prompt}. For brevity, the figure does not explicitly depict this part, and further details can be referred to in their work.}
   \label{fig:foreground harmonization}
   \vspace{-3ex}
\end{figure*}

\subsection{Imaging Condition Description Generation}

Recent breakthroughs in large vision-language models, such as Google Gemini and OpenAI GPT-4Vision, have paved the way for significant advancements. Gemini, trained on a massive multi-modality corpus encompassing vision and language, serves as an effective tool for providing descriptions for input composite images, making it an ideal choice for our work.

Our goal with Gemini is twofold. First, we aim to identify the foreground object, such as \textit{dog, flower, etc}. Second, we seek descriptions of the imaging conditions for both the foreground and background. Given potential differences in training language corpora between vision-language models (VLM) and text-to-image (T2I) models, some descriptions generated by the VLM may not be accurately interpreted by the T2I model. To address this, we provide a task specification prompt to the VLM, constraining the output scope with environmental text, including \textit{weather, season, time, color tone, etc}.

The input and output of the VLM can be defined as follows:
\begin{equation}
    [p_{Obj}, p_{ForeCond}, p_{BackCond}]=\mathrm{VLM}(S, M, I)
\end{equation}
Here, $S$ represents the task specification prompt, providing an instruction like \textit{``Please describe the imaging condition of both the foreground object and background"}. Additional details can be found in Appendix B. $M$ and $I$ denote the foreground mask and composite image, respectively. $[p_{Obj}, p_{ForeCond}, p_{BackCond}]$ represent the prompts for the object name, foreground environmental descriptions, and background environmental descriptions, ensuring one or more than one word. An example output could be: [\textit{``bird", ``bright orange", ``sunset blue"}].

\subsection{Foreground Region Harmonization}

In our context, the term "human prior" specifically refers to the inherent understanding that humans possess regarding the distribution of natural images. Since our daily experiences predominantly involve realistic and natural scenes, we intuitively internalize this distribution. Similarly, generative models, such as Stable Diffusion \cite{rombach2022high}, are trained on extensive datasets, enabling them to encode knowledge about the natural image distribution in their model weights \cite{bommasani2021opportunities}. Recognizing these shared characteristics, particularly in modeling the distribution of natural images, we establish a connection between generative models and the human prior and leverage Stable Diffusion in our work.

\textbf{Foreground Editing.} We employ the recent Prompt-to-Prompt (P2P) editing technology \cite{hertz2022prompt} to edit the foreground region. Capitalizing on the identified strong correlations between image pixels and text tokens, P2P achieves local image editing by maintaining fixed cross-attention maps while altering prompt texts. For instance, when transforming a \textit{dog} into a \textit{cat}, P2P preserves the original cross-attention map for the \textit{dog} word and simply replaces the guided word with \textit{cat}.

Inspired by P2P, a straightforward approach for image harmonization involves the following steps: Given an input composite image with descriptions of the imaging conditions for both the foreground and background, we initially use DDIM inversion \cite{song2020denoising} to project the image into the latent space for P2P utilization. Subsequently, we fix the cross-attention maps for the foreground description and change the guided words from foreground description to background one. The expectation is that the background's imaging condition description can guide the editing of the foreground region, thereby achieving image harmonization.

However, there exist two challenges in the above approach. Firstly, the provided descriptions of imaging conditions may not accurately represent the foreground and background. For images with rich elements, accurately describing them can be challenging, potentially misleading the T2I model's editing direction. The second issue is content distortion. P2P, initially designed for modifying object semantics (\textit{e.g.}, changing \textit{dog} to \textit{cat}), does not suit image harmonization well, which focuses on changing color tones without altering object context. Implementing the straightforward solution will result in nontrivial distortions in the foreground content structure.

To address these challenges, we propose an embedding refinement module and a structure preservation module (depicted in Fig. \ref{fig:foreground harmonization}). These modules optimize description text embeddings for accurate imaging condition representation and leverage self-attention maps and edge detection algorithms to preserve content structure.

\textbf{Text Embedding Refinement.} In the text-to-image process, Stable Diffusion utilizes text guidance by projecting words into higher-dimensional embeddings. While representing the environment at the word level may pose challenges, finding an accurate high-dimensional representative embedding is more feasible. Drawing from the observed strong correlations between embeddings and image pixels \cite{hertz2022prompt}, we introduce a quantitative indicator based on cross-attention maps to assess the representativeness of a text embedding in describing the environment: A good description of the imaging condition is characterized by well-focused cross-attention on the target region without overflowing or underfilling. Based on this indicator and inspired by textual inversion technique \cite{gal2022image}, we formulate the optimization objective to refine the embeddings of imaging condition descriptions:
\begin{equation}
\label{eq:Text optimizing}
\mathcal{L}_{Emb} = \Vert M - \frac{Att(Emb)}{max(Att(Emb))} \Vert_2^2
\end{equation}
Here, $Emb$ represents the optimized text embeddings. For multiple embeddings, $Att(Emb)$ is the summation of their cross-attention maps ($Att(Emb)=\sum^N_{n=1}Att(Emb_n)$), $M$ is the background mask, and $Att$ is the cross-attention map between the embedding and image pixels. To align with the value range of the mask, we normalize the attention map with its maximum value. In practice, we utilize the VLM to coarsely describe the imaging condition with some words and then minimize Eq. \ref{eq:Text optimizing} to obtain a more representative refined embedding. To prevent overfitting, we add a regularization term to ensure the refined embedding is close to its initial state:
\begin{equation}
\label{eq:Final Text optimizing}
\min_{Emb} (\mathcal{L}_{Emb} + w \Vert Emb - Emb_{init} \Vert_2^2)
\end{equation}
Here, $Emb_{init}$ is the embedding of the initial text, and $w$ is a hyper-parameter balancing the two terms.

Upon optimizing the text embeddings, we freeze the cross-attention maps of the foreground descriptions and substitute their embeddings with those from the background. However, since there may be a mismatch in the number of embeddings for foreground and background (\textit{e.g.}, 2 for foreground but 3 for background), and different background embeddings may have varying impacts, we replace each foreground embedding with a fusion of background embeddings:
\begin{equation}
    Emb_{F}=\alpha_1 Emb_{1}+\alpha_2 Emb_{2}+...+\alpha_N Emb_{N}
\end{equation}
Here, $\alpha_1+\alpha_2+...+\alpha_N=1$. The weights $\{\alpha_n\}^N$ are learned during the optimization process of the background descriptions by setting $Att(Emb)$ in Eq. \ref{eq:Final Text optimizing} to $\sum^N_{n=1}\alpha_nAtt(Emb_n)$. This approach allows the network to determine the importance of each word embedding.

\textbf{Content Structure Preservation.} Directly applying P2P editing technology with the optimized text embeddings may lead to content distortion. To harmonize the image while preserving content structure, we focus on two key aspects.

\textit{1) Self-attention exploitation}. Although P2P \cite{hertz2022prompt} has been successful in the local editing of images, we found that even if we changed the text of the environment but not the object itself, the content structure would distort a lot. In contrast to P2P, which relies solely on cross-attention, our approach incorporates additional self-attention maps. Previous studies \cite{shechtman2007matching, tumanyan2022splicing} have shown that self-similarity-based descriptors can capture structure while disregarding appearance information. By initially fixing the self-attention maps in Stable Diffusion for the foreground environment text, we achieve improved content retention. Leveraging the self-attention mechanism, each iteration of editing is effectively constrained to small changes (see Fig. \ref{fig:Pipeline}), mirroring human behavior.

\textit{2) Edge Detection Constraint.} Given that the self-attention map constraint provides a broad high-level constraint but may not ensure the preservation of finer details, we take an additional step to incorporate fine-grained edge constraints. Specifically, we utilize both a deep edge detection model (Pidinet) \cite{su2021pixel} and traditional Sobel kernels \cite{kanopoulos1988design} to extract edge maps:
\begin{equation}
    E_S(I)=(I\otimes K_h+I\otimes K_v) \quad\quad E_D(I)=\mathrm{Pidinet}(I)
\end{equation}
Here, $E_S$ and $E_D$ represent the Sobel edge map and deep edge map, respectively. $K_h$ and $K_v$ denote horizontal and vertical Sobel kernels, and $\otimes$ signifies convolution.

We implement the constraint by optimizing the unconditional (null-text) embedding of the Stable Diffusion. For each diffusion step, we decode the latent to the image and calculate the loss between the original image's edge and the resulting image's edge:
\begin{equation}
    \mathcal{L}_{Edge}=\Vert E_S(I) - E_S(I') \Vert_2^2+\gamma\Vert E_D(I) - E_D(I') \Vert_2^2
\end{equation}
Here, $I'$ denotes the decoded image in the diffusion steps, and $\gamma$ is a loss weight for balance. For more detailed information on null-text optimization, please refer to \cite{mokady2022null}.

\begin{figure}[t]
  \centering
   \includegraphics[width=\linewidth]{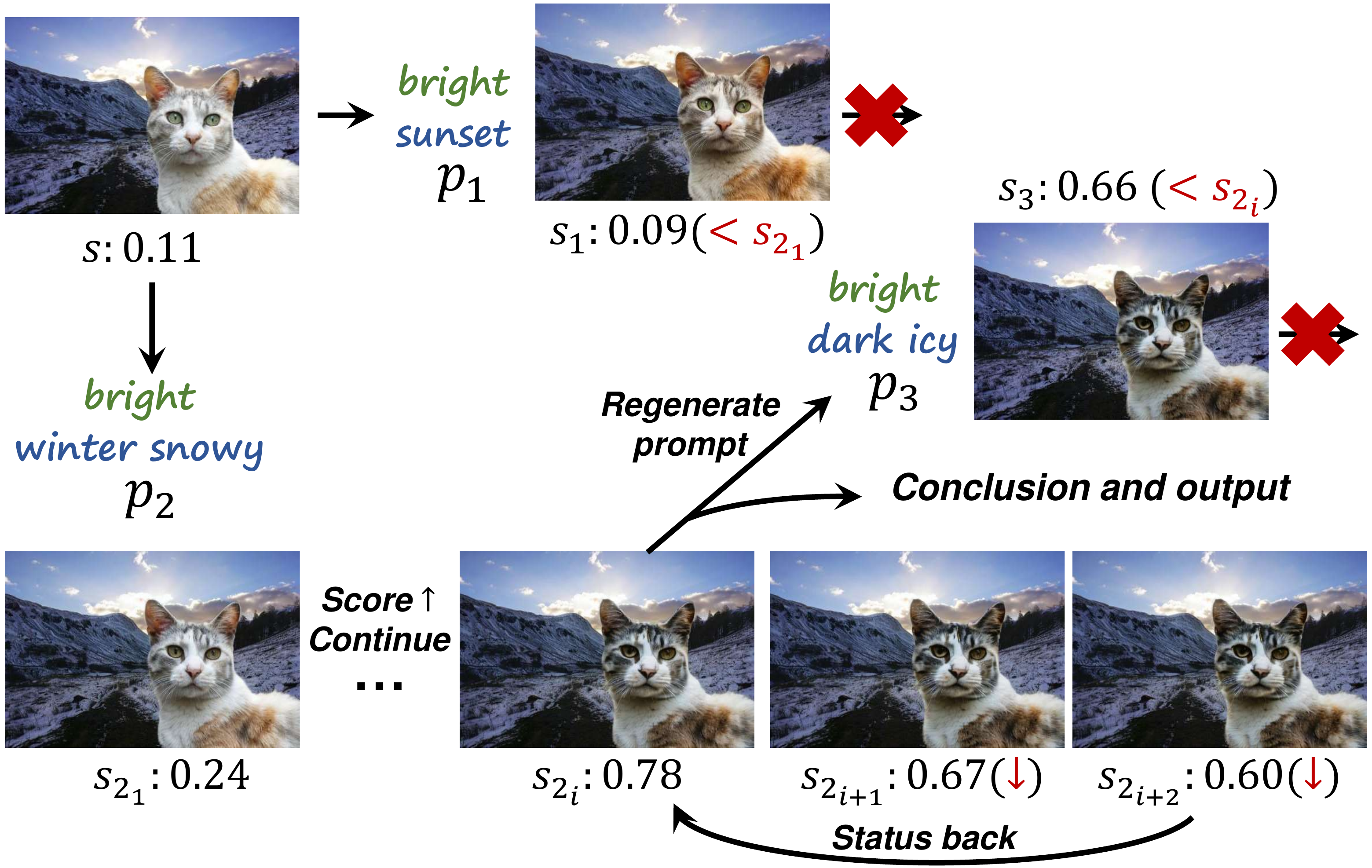}
   \vspace{-4ex}
   \caption{Demonstration of the evaluation process. We visualize how the decision of Continue/Regenerate/Conclude is made. Please zoom in for a better view.}
   \label{fig:evaluation_process}
   \vspace{-4ex}
\end{figure}

\begin{figure*}[t]
  \centering
   \includegraphics[width=0.95\linewidth]{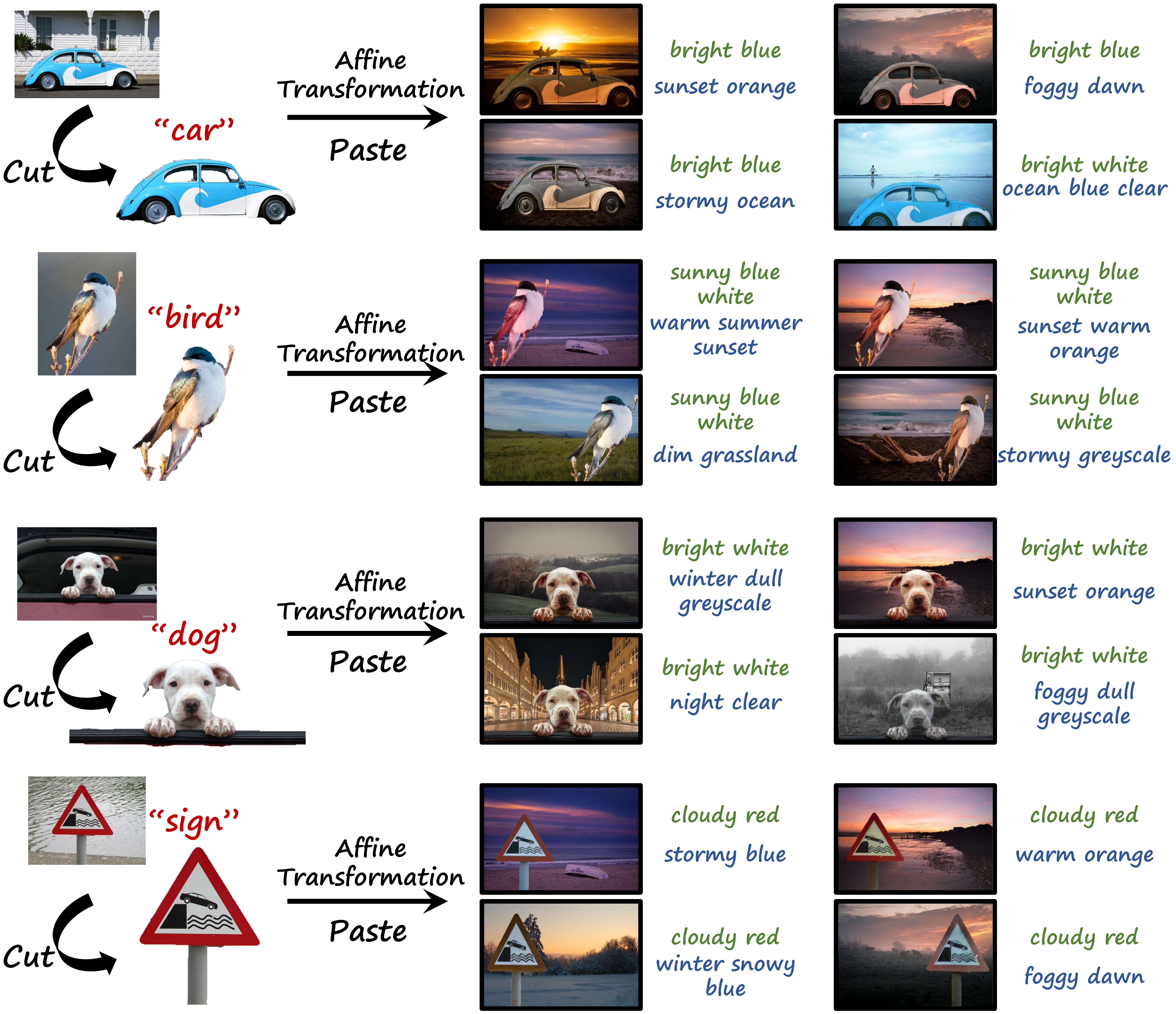}
   \vspace{-1.5em}
   \caption{Qualitative demonstration of our approach's harmonized results. In the leftmost column, the foreground source image and its extracted foreground object are depicted. The subsequent columns display the harmonized results of the foreground object under various background imaging conditions. The color scheme indicates the object name, descriptions of the foreground imaging condition, and descriptions of the background imaging conditions with \textbf{\textcolor{red}{RED}}, \textbf{\textcolor{green}{GREEN}}, and \textbf{\textcolor{blue}{BLUE}}, respectively. Note that multiple prompts may be leveraged in the process, and only the initially chosen description is displayed here.}
   \vspace{-1em}
   \label{fig:Demon1}
\end{figure*}

\begin{figure*}[t]
  \centering
   \includegraphics[width=0.95\textwidth]{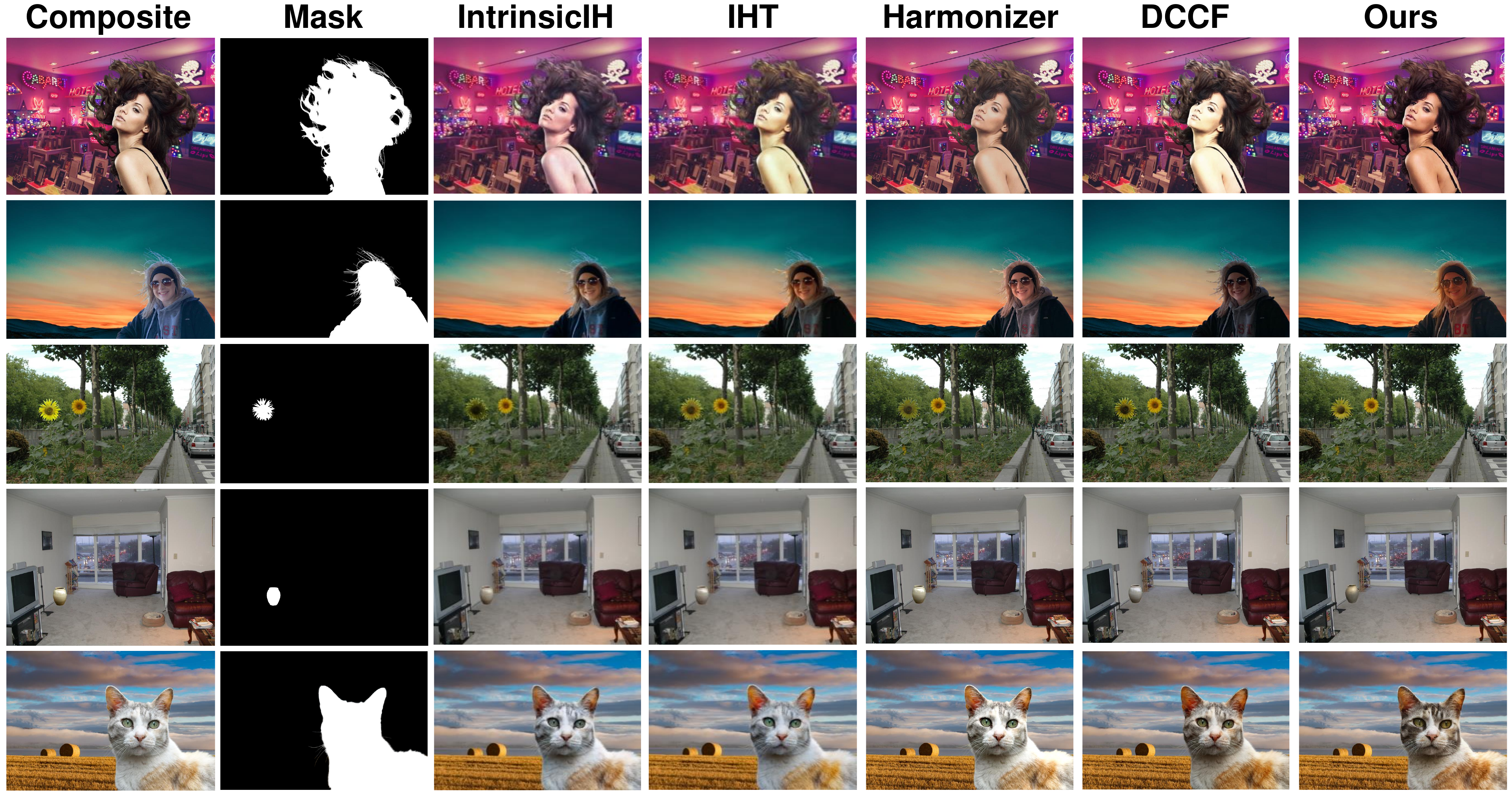}
   \vspace{-1em}
   \caption{Comparisons with SOTA harmonization methods. From left to right are composite images, foreground masks, and the harmonized results of Intrinsic \cite{guo2021intrinsic}, IHT \cite{guo2022transformer}, Harmonizer \cite{ke2022harmonizer}, DCCF \cite{xue2022dccf}, and ours.}
   \label{fig:Compares}
\end{figure*}

\begin{figure*}[t]
  \centering
   \includegraphics[width=0.85\linewidth]{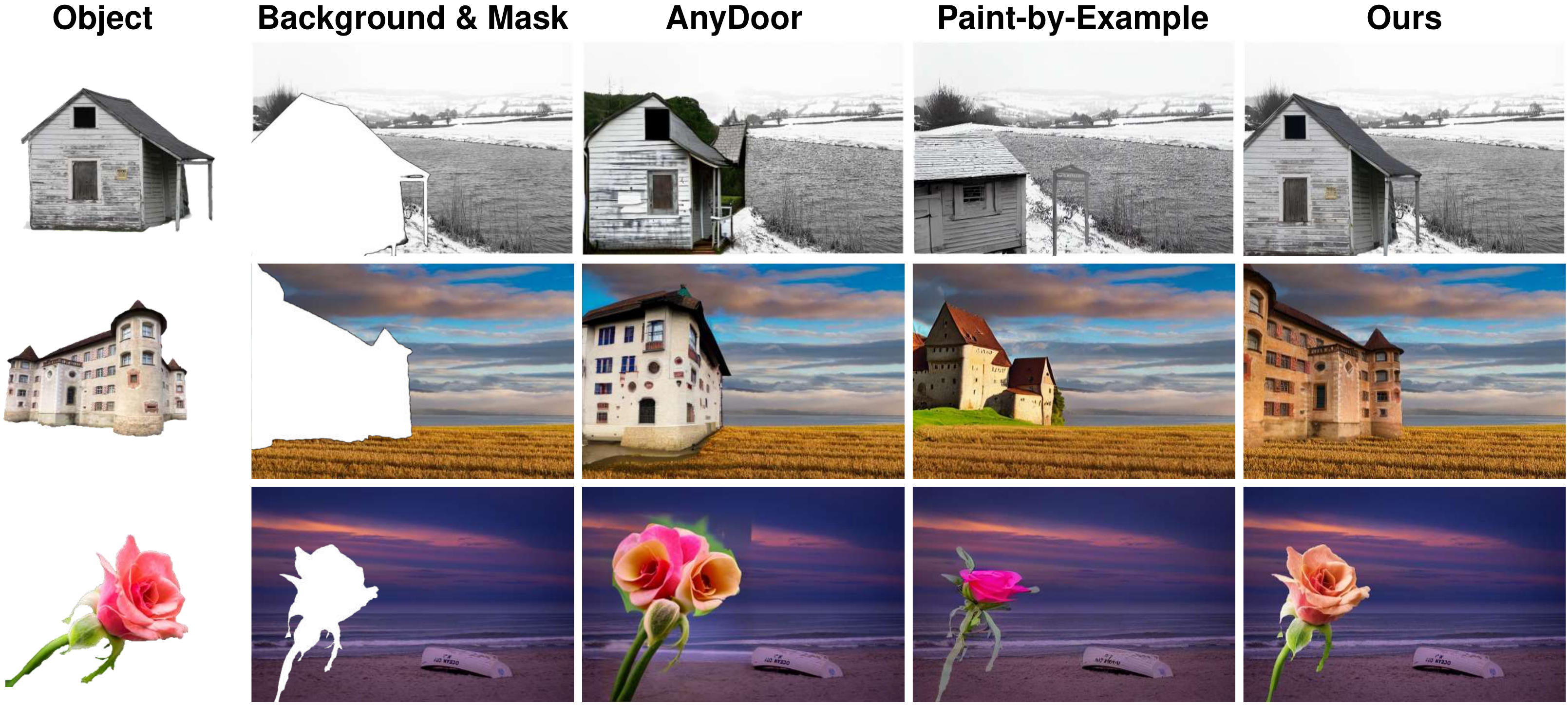}
   \vspace{-2ex}
   \caption{Comparisons with SOTA diffusion-based image composition methods. Please zoom in for a better view.}
   \label{fig:compare_with_composite}
   \vspace{-2ex}
\end{figure*}

\subsection{Performance Evaluation}

In the human harmonization process, encountering situations where the result is worse than the initial one after editing is not uncommon. In such cases, a human editor would reassess the evaluate the naturalness of the current result and editing direction.

To replicate this behavior and automate the entire pipeline further, we introduce a two-class classifier to assess the harmonization level of the rendered result. It's worth noting that this classifier serves as a post-processing step and does not alter the zero-shot nature of the previous harmonization process. Moreover, the classifier incurs minimal computational cost, utilizing significantly fewer images for training compared to previous supervised methods. For instance, training can be completed in just 30 minutes, whereas previous supervised methods may take over 2 days. More detailed information can be found in Appendix C.

The evaluation process is depicted in Fig. \ref{fig:evaluation_process}. At the beginning of harmonization, we use the VLM to generate multiple imaging condition descriptions, denoted as $p_1, p_2, ..., p_K$. Each description guides one harmonization iteration, and we pass the results into the classifier, obtaining scores $s_1, s_2, ..., s_K$. We select the prompt $p_k$ with the highest score and consistently use it in subsequent iterations. During intermediate harmonization iterations, if the score decreases twice in a row ($s_{k_{i+2}}< s_{k_{i+1}}< s_{k_{i}}$), we revert to the $s_{k_{i}}$ state and regenerate the description of the imaging condition to explore whether the new description can guide us to a higher harmonization score. If successful, we continue the harmonization process; otherwise, we conclude the process and output the result with the highest harmonization score as the final outcome.

\section{Experiments}

Here, we present a qualitative demonstration of our method and compare it with supervised harmonization methods. Additional details regarding the implementation, ablation study, and potential applications can be found in Appendix C, D, and E.

\subsection{Dataset Setup}

To evaluate the effectiveness of our approach in harmonizing real-world composite images, we engaged two image editors to help construct a set of composite images for assessment. Our process involved collecting high-quality images from Flickr, using some as source images for foreground objects and others as source images for the background environment. Various objects, such as humans, animals, buildings, \textit{etc.}, were extracted from the foreground images and seamlessly integrated into suitable positions in the background image after undergoing transformations like resizing, rotation, and perspective adjustments. In total, we compiled a dataset of 300 composite images for comprehensive evaluation.

\subsection{Qualitative Demonstration}

\begin{figure*}[t]
  \centering
   \includegraphics[width=\linewidth]{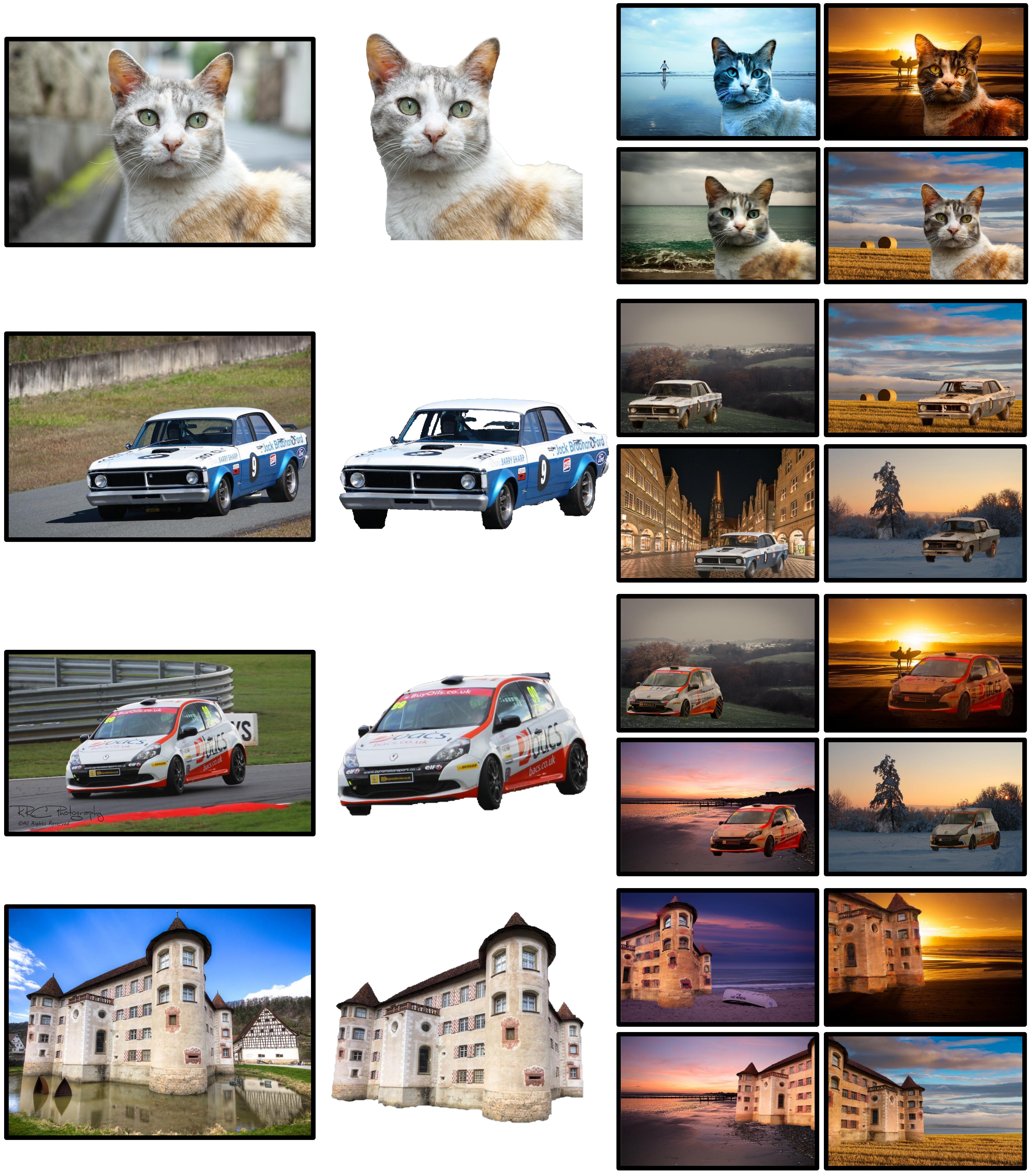}
   \caption{Additional qualitative demonstration of our approach's harmonized results. In the left two columns, the foreground source image and its extracted foreground object are depicted. The subsequent columns display the harmonized results of the foreground object under various background imaging conditions.}
   \label{fig:Demon2}
\end{figure*}

We provide qualitative examples of harmonized results across different objects and imaging conditions in Fig. \ref{fig:Preface}, Fig. \ref{fig:Demon1}, and Fig. \ref{fig:Demon2}. These results showcase the versatility of our approach in effectively handling a variety of real-world composite images. In Fig. \ref{fig:Demon1}, it is evident that for the same background image, different descriptions are chosen for different foreground objects, illustrating an adaptive process for selecting more suitable descriptions. These examples highlight the robustness of our approach.

\subsection{Human Preference Evaluation}
\label{subsec:Comparisons}

To demonstrate the effectiveness of our approach, we compare it against several popular heavily supervised harmonization methods, including IHT \cite{guo2022transformer}, DCCF \cite{xue2022dccf}, Harmonizer \cite{ke2022harmonizer}, and IntrinsicIH \cite{guo2021intrinsic}. The qualitative results are presented in Fig. \ref{fig:Compares}, where our method, even without relying on heavy training, achieves superior performance to other methods, highlighting the effectiveness of our zero-shot approach.

Additionally, we compare our approach with recent works that leverage diffusion models for image composition tasks. Given their inherent relation to image harmonization, we visualize comparisons with popular methods like AnyDoor \cite{chen2023anydoor} and Paint-by-Example \cite{yang2023paint}. From the results in Fig. \ref{fig:compare_with_composite}, it's evident that these methods fail to preserve the foreground object's structure, while ours performs better in this aspect.

For quantitative comparison, as ground-truth harmonized results are unavailable for these real-world composite images, we follow the approach in \cite{gal2022image, mokady2022null, avrahami2022blended, wu2022uncovering, xie2022smartbrush} and conduct a subjective user study. We invited 40 image editors with more than 2 years of work experience for the study. In each trial, participants were presented with two images, which were either derived from the results of various harmonization methods or represented the original composite image. The display order was randomly shuffled, and participants were required to choose the more harmonious one between the two. A total of 60,000 votes were collected. The vote rates are displayed in Table \ref{fig:UserStudy}, demonstrating the effectiveness and superiority of our approach.


\begin{table}[t]
\caption{User study. Users are invited to select the more harmonious one between two images. Here, we display the voting percentages of different methods \cite{xue2022dccf, ke2022harmonizer, guo2021intrinsic, guo2022transformer}. ``Comp." denotes the original composite image.}
\resizebox{0.7\linewidth}{!}{\begin{tabular}{l|c}
\hline
Comparisons           & Voting Rate   \\ \hline
Ours \textit{v.s.}  Comp.      & \textbf{89.6\%}/10.4\% \\
Ours \textit{v.s.} IntrinsicIH & \textbf{62.6\%}/37.4\% \\
Ours \textit{v.s.} IHT         & \textbf{60.3\%}/39.7\% \\
Ours \textit{v.s.} Harmonizer  & \textbf{72.1\%}/27.9\% \\
Ours \textit{v.s.} DCCF        & \textbf{70.9\%}/39.1\% \\ \hline
\end{tabular}}
   \label{fig:UserStudy}
\end{table}

\section{Limitation}
\label{Apd:Limitations}

While our approach offers a modularized framework, it may encounter accumulated errors, such as potential inaccuracies in description generation, deviations in text-to-image editing directions, or inaccuracies in the evaluation classifier. However, this modular design also provides ample room for future improvement. As these networks advance, our techniques should also see improvements due to their modular nature. Another limitation is the time cost associated with our approach. As the diffusion model involves an iterative generation process, our method inherits the drawback of increased time consumption. Additionally, due to the fundamental limitations of the T2I model, despite several designs for structure preservation, there are still deformations in the foreground region, especially in finer details.

\section{Conclusions}

In this work, we introduce the feasibility of imitating human behavior and propose a zero-shot image harmonization method. Our approach achieves satisfactory harmonized results without relying on extensive training on a large dataset of composite images. Instead, it leverages the prior knowledge embedded in the pretrained generative model and the guidance of textual descriptions describing the image environment. The effectiveness of our method is demonstrated across a variety of cases. With a fully modularized design, our framework allows for easy integration of future improvements, leveraging the rapid advancements in vision and language models. We hope that our method serves as a stepping stone for further research and development in the field of zero-shot image harmonization.

\bibliographystyle{ACM-Reference-Format}
\bibliography{sample-bibliography}

\clearpage

\appendix

\section{Overview}
\label{Apd:Overview}
In this supplementary file, we will begin by presenting the task specification prompt in Appendix \ref{Apd:prompt}. Following that, we will delve into the implementation details of our proposed method in Appendix \ref{Apd:Implementation Details}. Subsequently, in Appendix \ref{apd:ablation}, we provide an in-depth exploration of ablation studies. Finally, we will investigate some practical applications of our method in Appendix \ref{apd:applications}.

\begin{figure*}[!h]
  \centering
   \includegraphics[width=\linewidth]{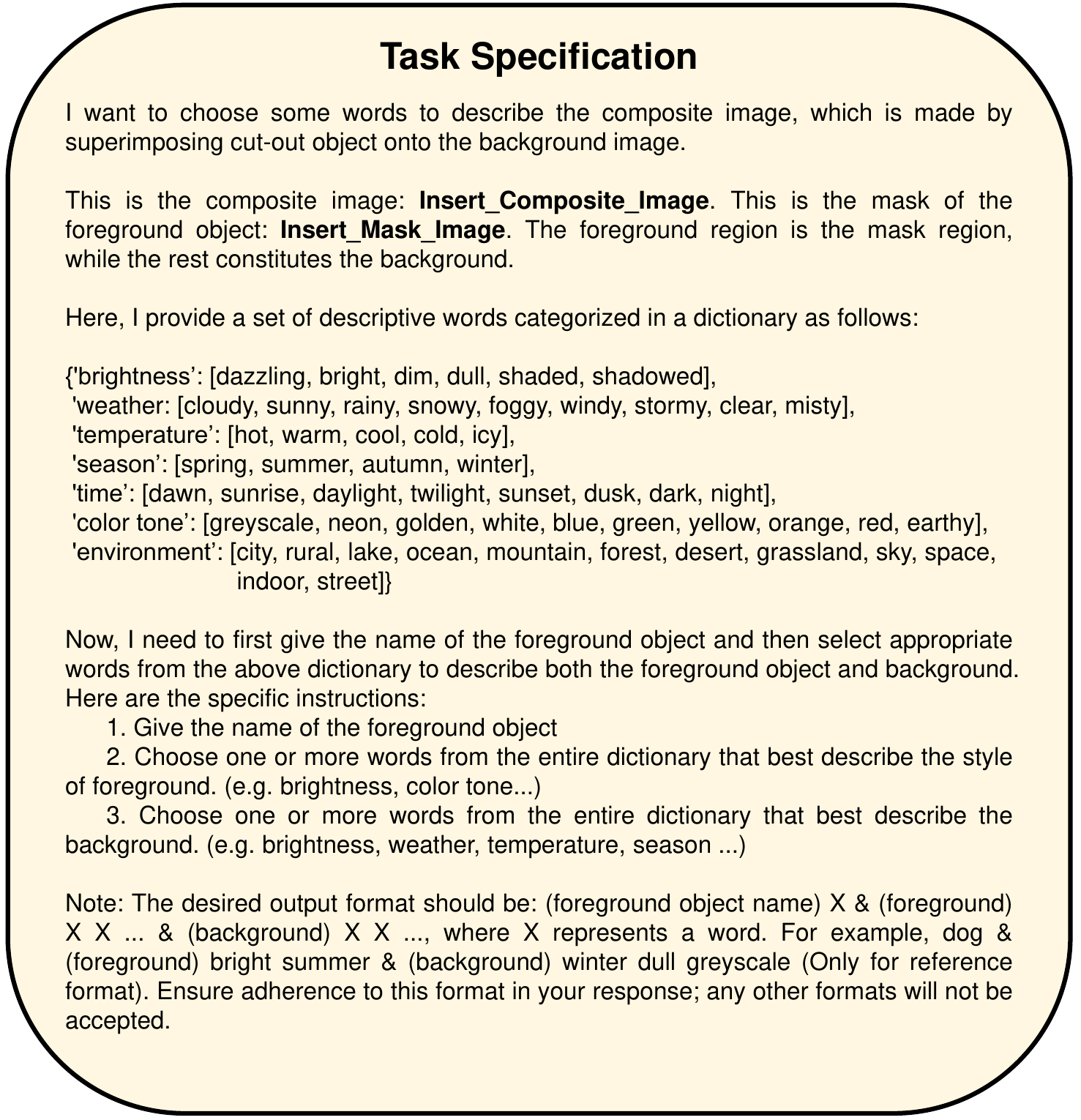}
   \caption{Task specification used for generating descriptions of object names, foreground imaging conditions, and background imaging conditions employing Google Gemini.}
   \label{fig:task specification}
\end{figure*}

\section{Task Specification Prompt}
\label{Apd:prompt}

In Fig. \ref{fig:task specification}, the complete prompt for task specification is presented, guiding the Vision-Language Model (VLM) to produce satisfactory results. To address potential disparities in language corpora between the VLM and the T2I model, we encompass the scope of descriptions related to the imaging condition, ensuring easy interpretation by the T2I model.

\section{Implementation Details}
\label{Apd:Implementation Details}

In our approach, we leverage Google Gemini as the VLM. At the beginning of harmonization, we will generate 3 different imaging condition descriptions. We adopt the Stable Diffusion model \cite{rombach2022high} as the Text-to-Image (T2I) model. The input resolution is set to $512 \times 512$. We utilize the DDIM sampling schedule \cite{song2020denoising} for a deterministic diffusion process, configuring the number of diffusion timesteps to 50. In DDIM inversion, we eliminate conditional guidance by setting the guidance scale to 0. Conversely, in the reverse process, the guidance scale in the classifier-free guidance \cite{ho2021classifier} is set to 2.5.

The refinement of text embedding is based on the cross-attention maps in the diffusion process. We explore two implementations: optimizing style and training style. The former optimizes the text embedding in each diffusion timestep, similar to \cite{mokady2022null}. In each step, we optimize the embedding multiple times based on the embedding from the last step, obtaining a series of optimized embeddings by the final timestep. In contrast, the latter trains a single embedding. Using all timesteps as a dataset, we iteratively sample a batch to train the embedding. Both implementations yield satisfactory harmonized results, with the optimizing style being the default unless specified. When adopting AdamW \cite{loshchilov2017decoupled} with an initial learning rate of $1e^{-3}$ ($1e^{-2}$ for training style) and setting $w$ in Eq. 3 in the main paper to 5000 (1000 for training style), we optimize two times in each timestep (training for 50 epochs with a batch size of 4). We found that both implementations can achieve satisfactory harmonized results. If not specified, we leverage the optimizing style in the context. 

Regarding Pidinet and Sobel kernels, we observe that Pidinet may fail to capture detailed edges compared to Sobel kernels. Therefore, we set the weight $\gamma$ in Eq. 6 in the main paper as 0.1.

For the performance evaluator, we utilize the ResNet-50 structure. In its training, we use 1000 generated results and invite 2 image editors to score them on a scale from 0 to 1 in 10 ranks (\textit{i.e.}, 0.1, 0.2, ..., 1). These scores serve as soft labels to train a two-class classifier. All experiments are conducted on a single RTX 4090 GPU.

\begin{figure}[t]
  \centering
   \includegraphics[width=\linewidth]{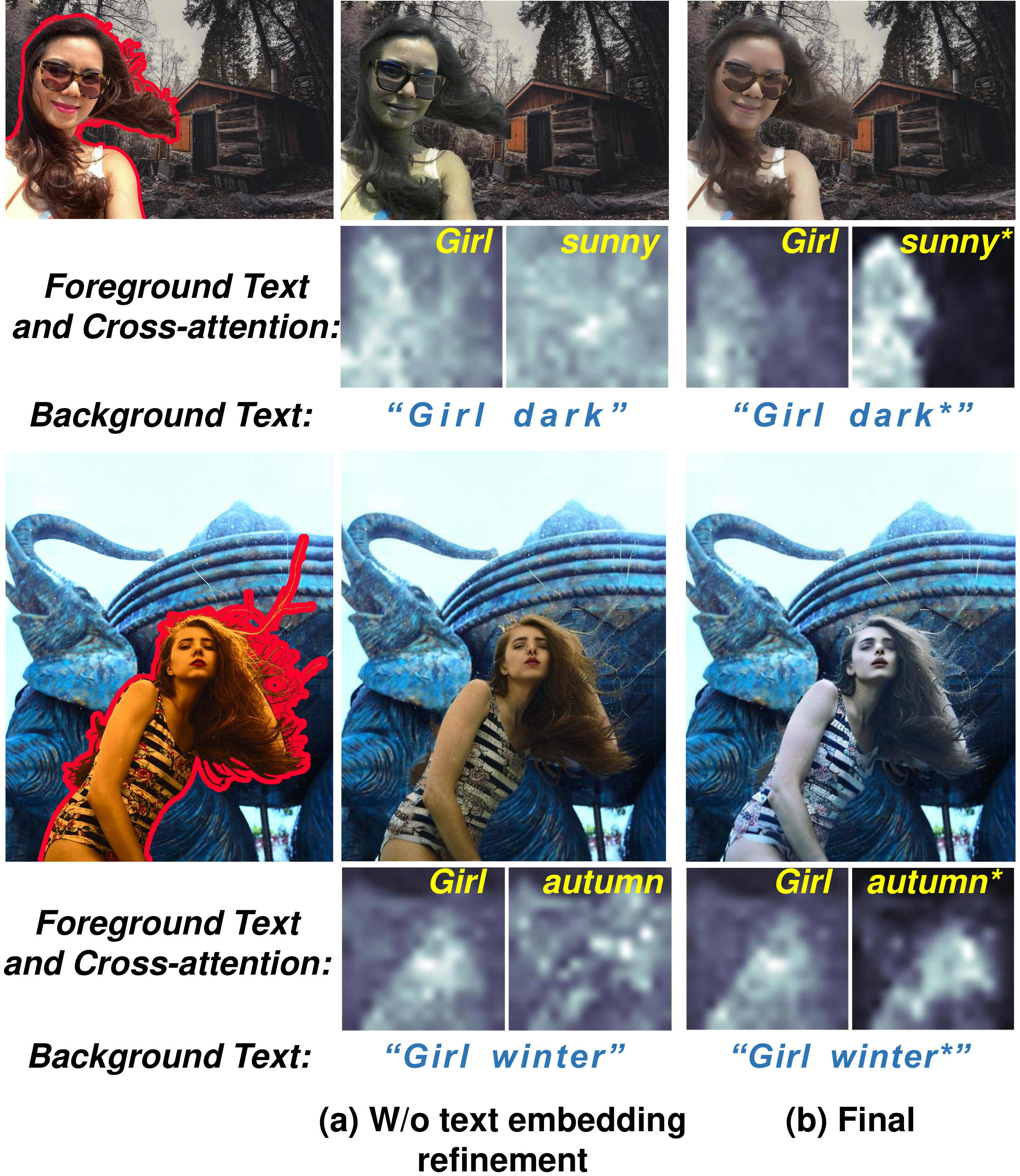}
   \vspace{-4ex}
   \caption{Ablation of the text embedding refinement.}
   \label{fig:AblationText}
\end{figure}

\section{Ablation Study}
\label{apd:ablation}

\textbf{Ablation of Text Embedding Refinement}. In Fig. \ref{fig:AblationText}, we demonstrate the effectiveness of the proposed Text Embedding Refinement module. Without optimization, it can be seen from the attention maps that the initial text fails to well illustrate the environments of foreground/background areas, resulting in a deviated harmonization direction.

\begin{figure}[t]
  \centering
   \includegraphics[width=\linewidth]{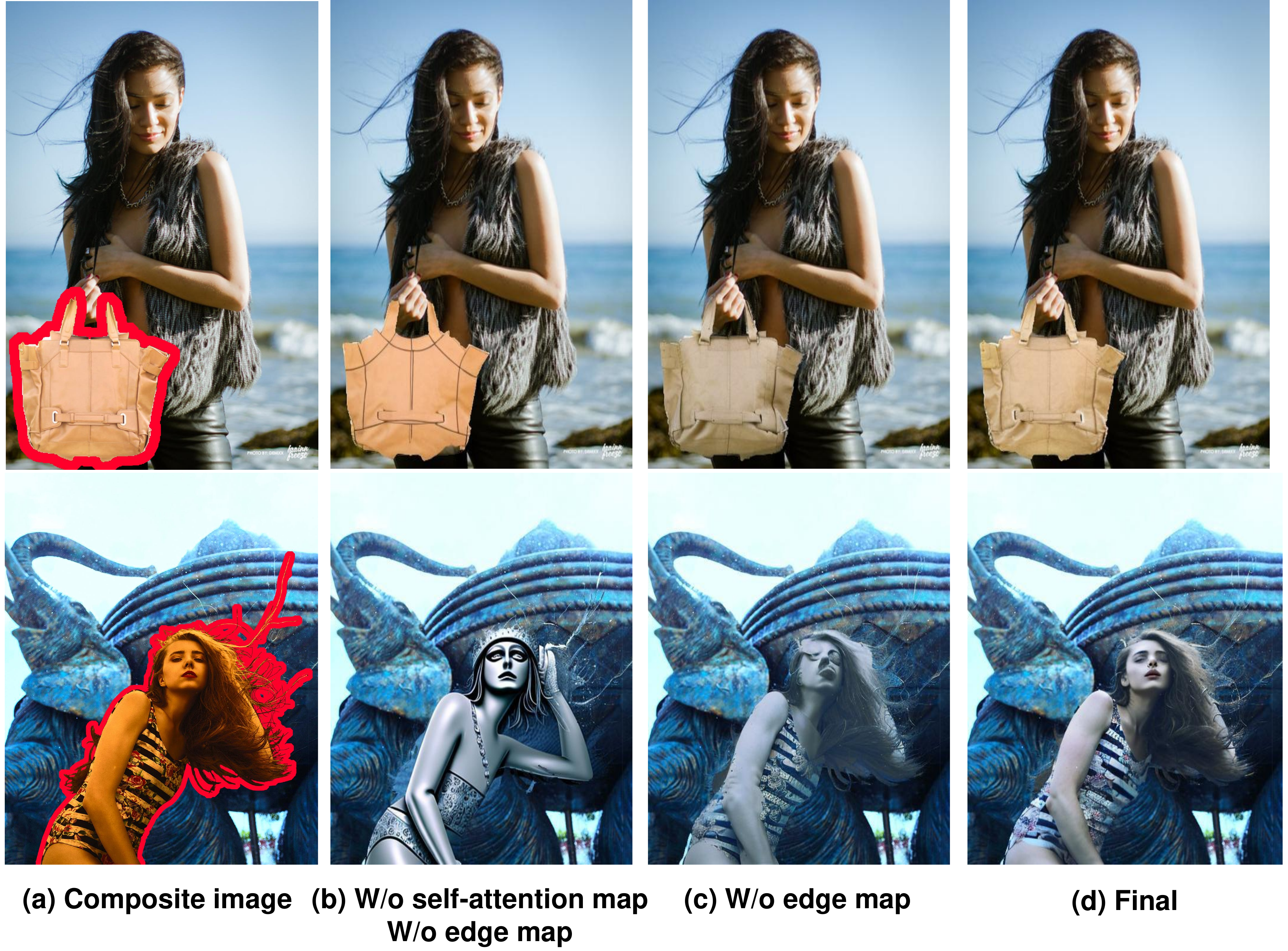}
   \vspace{-4ex}
   \caption{Ablation of the operations for content structure preservation. We ablate the utilization of self-attention maps and the edge maps here.}
   \label{fig:AblationContent}
   \vspace{-1ex}
\end{figure}

\textbf{Ablation of content retention designs}. From Fig. \ref{fig:AblationContent}, we can observe that with the self-attention and edge maps, the content structure is largely preserved.

\begin{figure}[t]
  \centering
   \includegraphics[width=\linewidth]{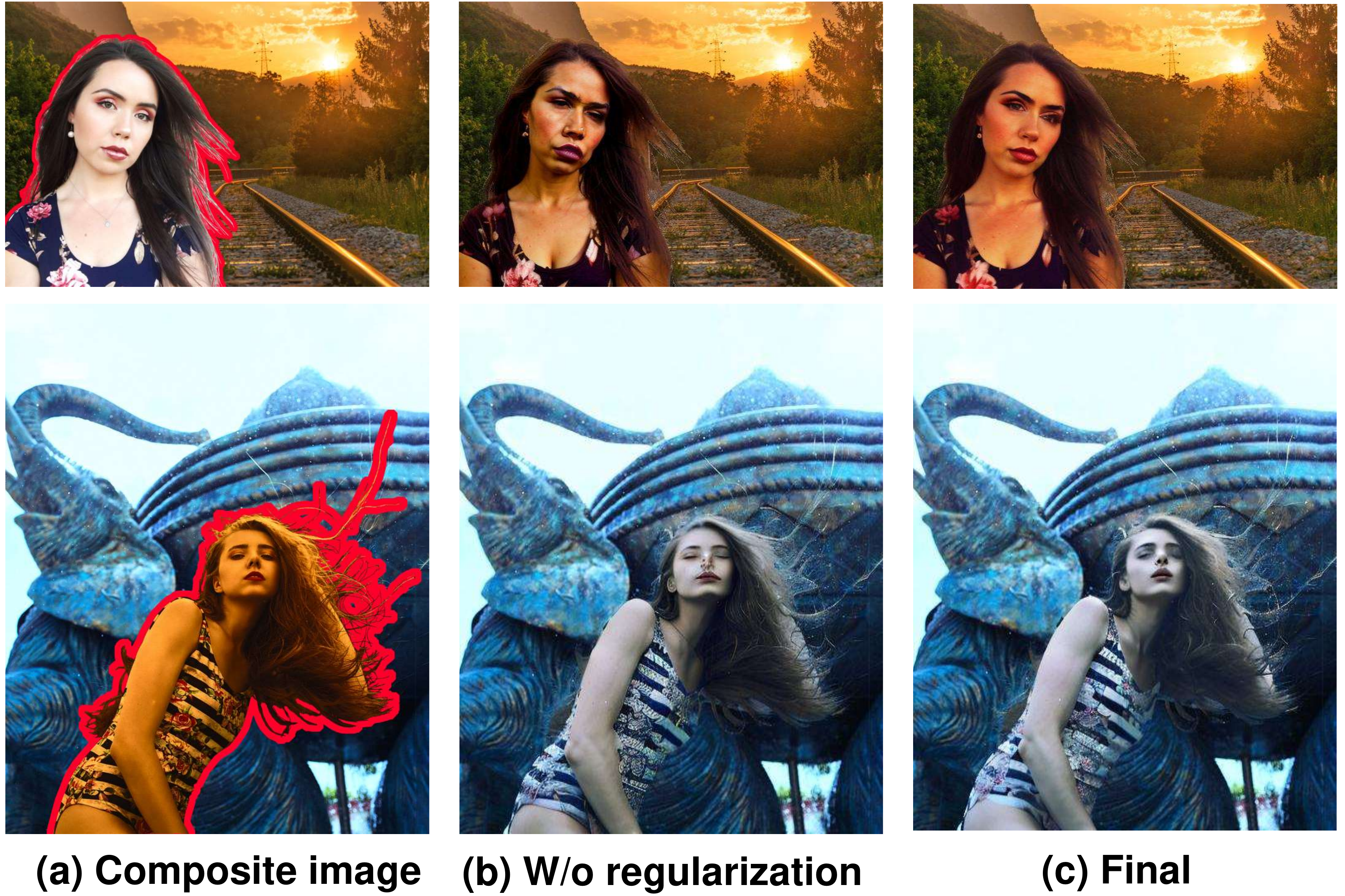}
   \caption{Ablation of the regularization term in the text embedding refinement.}
   \label{fig:AblationReg}
\end{figure}

\begin{figure}[t]
  \centering
   \includegraphics[width=\linewidth]{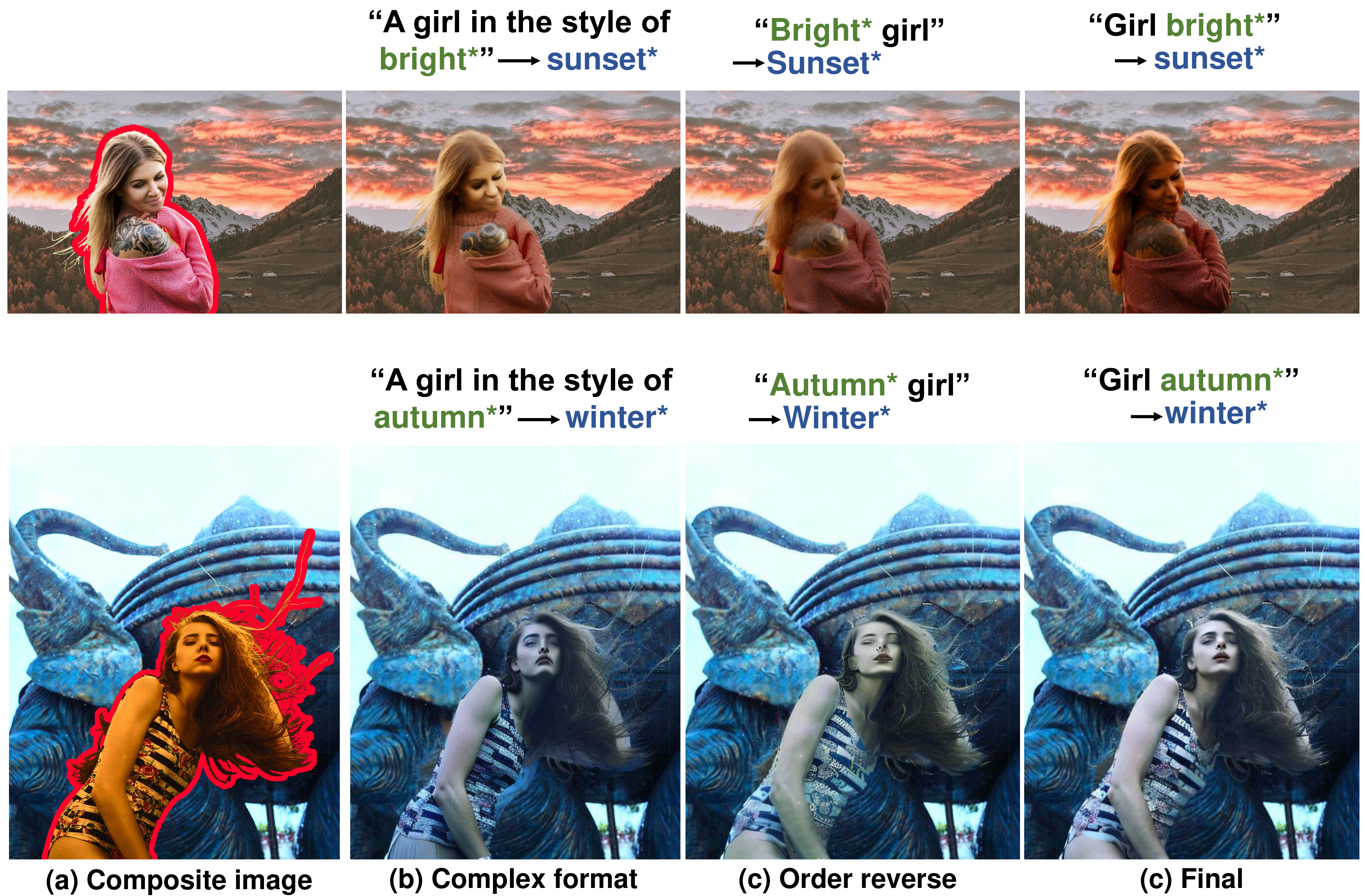}
   \caption{Ablation of the format and order of the initial text. The effect of formal/simple text is explored, together with whether the environment text is placed at the end. Please zoom in for a better view.}
   \label{fig:AblationPrompt}
\end{figure}

\textbf{Regularization of Text Embedding Refinement.}
In Eq. 3 of the main paper, we leverage an additional regularization term to restrict the optimization of text embedding. We ablate the regularization in Fig. \ref{fig:AblationReg}, and the results have demonstrated its effectiveness. With the regularization, the text embedding will not go far away from the initial and thus play a better guiding role.

\textbf{Prompt text format.}
Our text prompt is intentionally kept simple, consisting of only a noun for the object name and some adjectives for the imaging conditions. Interestingly, when examining examples from various large-scale text-to-image models \cite{ramesh2021zero, ding2021cogview, saharia2022photorealistic, ramesh2022hierarchical, yu2022scaling, rombach2022high}, it is observed that these models often leverage more complete and formal sentences to describe a given image. In an exploration of the impact of different prompt text formats, as illustrated in Fig. \ref{fig:AblationPrompt}, we observe that the currently used simplified texts lead to superior results. We attribute this outcome to the potential constraints introduced by longer prompt texts. The use of more extended prompts may introduce additional perturbations in the process, while the simplicity of our text format appears to facilitate the harmonization process significantly.

\textbf{Prompt text order.}
Interestingly, we observed that altering the order of prompt texts results in variations in the generated images. Specifically, when the environment text precedes the content text, there is a noticeable decrease in performance, as illustrated in Fig. \ref{fig:AblationPrompt}. Upon closer examination of the internal structure of the Stable Diffusion \cite{rombach2022high}, we identified the causal attention mask as a contributing factor. This attention mechanism blends the semantics of the latter text token with the preceding token. Consequently, placing the content text after the environment text makes it susceptible to the optimization process of the preceding environment text. This, in turn, leads to challenges in preserving the content structure. Notably, this observation aligns with the findings reported by \cite{feng2022training}.

\begin{figure}[t]
  \centering
   \includegraphics[width=\linewidth]{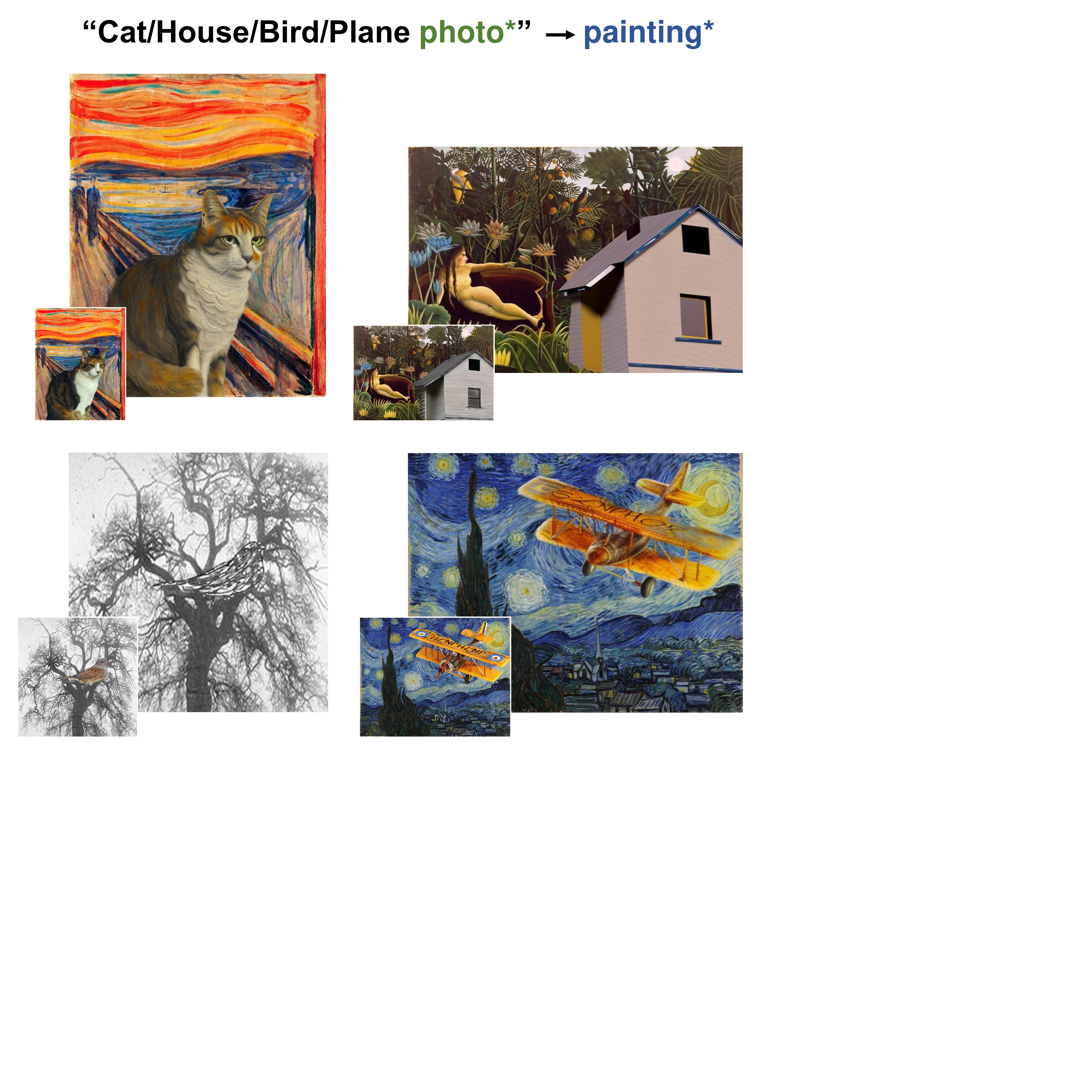}
   \caption{Further application of painterly image harmonization. We explore generalizing our method on artwork. The small image in the lower left corner of each image is the original composite image. Please zoom in for a better view.}
   \label{fig:ArtisticHarmon}
   \vspace{-2ex}
\end{figure}

\begin{figure}[t]
  \centering
   \includegraphics[width=\linewidth]{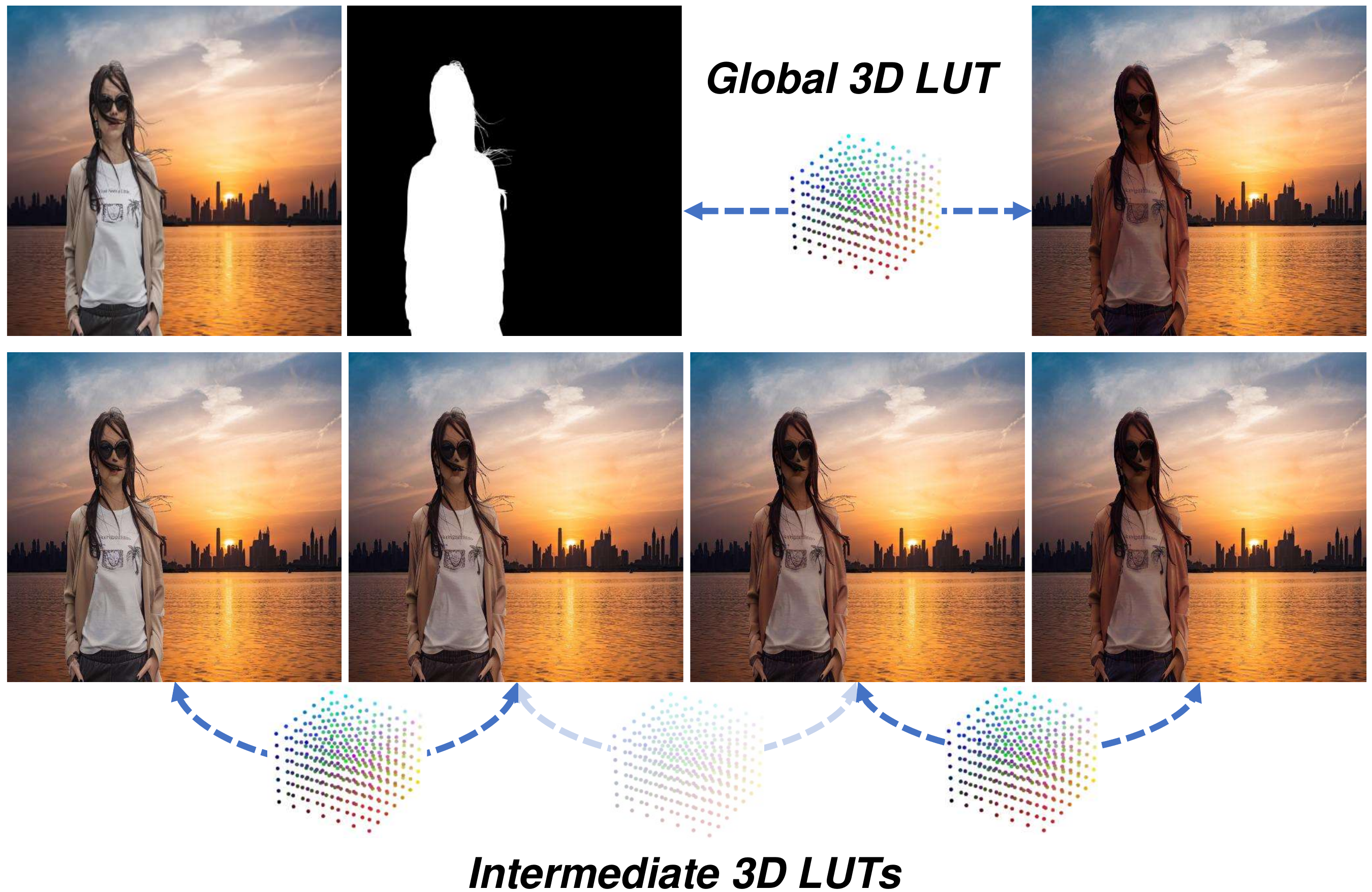}
   \caption{Further application as user auxiliary. With our iterative harmonization process, the intermediate operation of the network can be extracted and converted to 3D LUTs, which can be leveraged as guidance for the human process.}
   \label{fig:LUTGuidance}
   \vspace{-2ex}
\end{figure}

\section{Applications}
\label{apd:applications}

In this subsection, we go a further step to explore the applications of our method. We mainly focus on three: painterly image harmonization, user auxiliary, and multiple instances harmonization.

\textbf{Painterly image harmonization}. Besides the harmonized results on the natural composite images displayed in the main paper, we try to generalize our method to composite images with a large style gap between the foreground and background. Here, we evaluate the performance on some artistic pictures. To achieve better results, we slightly relax the constraint on content retention by utilizing self-attention maps only in certain diffusion timesteps (for natural images all timesteps are considered). From Fig. \ref{fig:ArtisticHarmon}, it can be seen that our method still succeeds in harmonizing the artwork images. We attribute this success to the powerful generative model prior. As mentioned in Sec. 1 of the main paper, a human mainly relies on his long-term prior on harmonious images to conduct the image harmonization process, and can often achieve harmonious results. Similar to that, thanks to the extremely large amount of real-world training data, the pretrained large-scale generative models inherently have that prior. Thus, our method, based on the generative model prior, can gain better generalization to a variety of image types, while the existing approaches \cite{guo2022transformer, cong2022high, xue2022dccf, hang2022scs, ke2022harmonizer, jiang2021ssh, chen2023dense} are limited to be applied to the specific scope of the synthesized composite images seen in their training phase.

\textbf{User auxiliary}. As depicted in Fig. 3 in the main paper, our method is an iterative harmonization process. Each time the composite image is harmonized a little. Thus, compared with the one-step methods \cite{hang2022scs, guo2022transformer, cong2020dovenet, tsai2017deep}, the user observes all the intermediate results, which can bring many benefits and leads the method to be a white-box approach. Specifically, due to the trivial content distortion between the results of two adjacent iterations, we can optimize a 3D LUT \cite{zeng2020learning, karaimer2016software}, which is a lookup table that maps RGB values between two images. By predicting either a global 3D LUT for the entire foreground region or multiple local LUTs for foreground meaningful parts such as faces, we can reveal the latent operations of the model, which in turn can provide guidance for an image editor to harmonize images. The process is illustrated in Fig. \ref{fig:LUTGuidance}.

\begin{figure}[t]
  \centering
   \includegraphics[width=\linewidth]{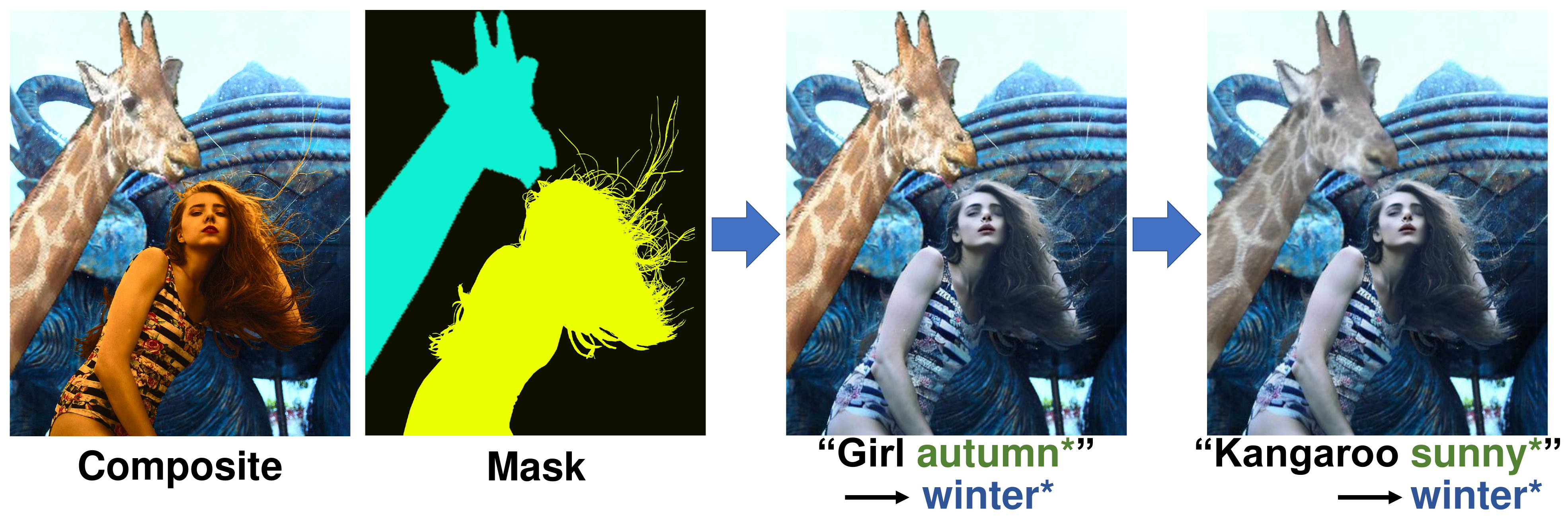}
   \caption{Pipeline for harmonizing multiple instances.}
   \label{fig:Multi-Instances}
\end{figure}

\textbf{Multiple instances harmonization}.
Since we need an initial environmental text for foreground/background, our method may not work well when processing multiple instances simultaneously. However, if we break it down into several separate parts, then we can make it (see Fig. \ref{fig:Multi-Instances}). This will not affect the user experience much, as in real-world scenarios, users mostly add foreground instances to the background image one by one. 

\end{document}